# Similarity Measures on Preference Structures, Part II: Utility Functions

**Vu Ha**★
★ Department of EE&CS
University of Wisconsin-Milwaukee
PO Box 784
Milwaukee, WI 53211
*vu@cs.uwm.edu*

**Peter Haddawy**★†
† CSIM Program
School of Advanced Technologies
Asian Institute of Technology
Bangkok, Thailand
*haddawy@cs.ait.ac.th*

**John Miyamoto**‡
‡ Department of Psychology
University of Washington
Box 351525
Seattle, WA 98195
*jmiyamot@u.washington.edu*

## Abstract

In previous work [8] we presented a case-based approach to eliciting and reasoning with preferences. A key issue in this approach is the definition of similarity between user preferences. We introduced the probabilistic distance as a measure of similarity on user preferences, and provided an algorithm to compute the distance between two partially specified *value* functions. This is for the case of decision making under *certainty*. In this paper we address the more challenging issue of computing the probabilistic distance in the case of decision making under *uncertainty*. We present algorithms to compute the probabilistic distance between two completely or partially specified *utility* functions. We demonstrate the use of this algorithm with a medical data set of partially specified patient preferences, where none of the other existing distance measures appear definable. Using this data set, we also demonstrate that the case-based approach to preference elicitation is applicable in domains with uncertainty.

## 1 INTRODUCTION

In previous work [8], we propose a case-based approach to preference elicitation. Assuming the existence of a population of users from whom we have elicited complete or incomplete preferences, we propose eliciting the preferences of a new user interactively and incrementally, using the closest existing preference structures as potential defaults. We envision our system to maintain a population of users with their preferences partially or completely specified in a given domain. When encountering a new user $A$, the system elicits some preference information from $A$ and then determines which user in the population has the preference structure that is closest to the preference structure of $A$. The preference structure of that user will be used to determine an initial default representation of $A$'s preferences.

This approach originates from the observation that people tend to form clusters according to their preferences or tastes, an observation that has been analyzed in the area of market segmentation [7]. It is also inspired by recent work on collaborative filtering [15], in which the filtering system predicts how interesting a user will find items he has not seen based on the ratings that other users give to items. Each user in a population rates various alternatives, e.g. newsgroup postings or movies, according to a numeric scale. The system then correlates the ratings in order to determine which users' ratings are most similar to each other. Finally, it predicts how well users will like new articles based on ratings from similar users.

One key issue common to this approach and the works in collaborative filtering is the choice of a distance measure on preference orders. In [8], we introduced a novel distance measures, called the *probabilistic distance*. According to this measure, the distance between two preference orders is determined by the probability that they disagree in their relative rankings of two randomly picked decision consequences. We provided an approximate algorithm to compute the probabilistic distance between two partial preference orders in the case of decision making under certainty. This work was later implemented in DIVA, a Decision-Theoretic Video Advisor that recommends movies [13]. Empirical analysis with DIVA showed that using the probabilistic distance results in more accurate recommendations than using the predominant Pearson's correlation measure.

In this paper, we tackle the outstanding issue of computing the probabilistic distance on preference orders in the case of decision making under uncertainty. We show that, under a reasonable assumption, this prob-lem is reduced to the well-studied problem of computing the volumes of convex bodies for which efficient, randomized algorithms have been developed. A key ingredient of these algorithms is a Markov chain-based, polynomial time sampling algorithm that samples points from a convex body according to a nearly uniform distribution. We propose to use this sampling algorithm directly to estimate the probabilistic distance on partially specified utility functions.

## 2  A MOTIVATING EXAMPLE

In the area of collaborative filtering, recommender systems such as GROUPLENS [15] and the DIVA video recommender [13] all require the use of a distance measure on preferences. Because all of these systems concern with decision making under *certainty*, it is not clear whether a study of distance measures on preferences is warranted in the case of *uncertainty*. We argue that it is. The concept of "how different is my preference from yours" is intuitive, but far from well-understood, especially when the preferential information is incomplete, or the choices are uncertain, or both. We shall now describe an example to illustrate this point.

Miyamoto and Eraker [12] described a psychology experiment with 44 undergraduate students at the University of Michigan. The experiment is designed to test several assumptions about people's preferences and attitudes towards risks with regards to survival duration. The subjects were asked to assign certainty equivalences (CE) to a total of 42 standard gamble questions involving duration of survival. Below is a typical question:

> *For any non-negative number $n$, let $n$ be the event that you will live exactly $n$ more years in good health, and then have a sudden and relatively painless death. Let $(m, .5, n), 0 \leq m < n$, be a lottery of 50% chance for $m$ and 50% chance for $n$. What is the number $p$ for which you regard $(m, .5, n)$ and $p$ as equivalent (denoted $(m, .5, n) \sim p$)?*

Suppose that $u$ denotes the utility function of a subject. Each answer of the form $(m, .5, n) \sim p$ translates into the following constraint on $u$: $u(m) + u(n) = 2u(p)$. Thus for each subject, we have a set of 42 constraints on his/her utility function $u$. Given two subjects with utility functions $u$ and $u'$, how should we define a distance measure between $u$ and $u'$? A simplistic approach may use some well-known statistical measures such as Spearman's footrule, Ulam distance, or various correlation coefficients. The problem with this approach is twofold. First, it typically requires that the constraints on $u$ and $u'$ are obtained from exactly the same set of CE questions, which substantially reduce its applicability. Second, this approach has to address sensitivity issues with respect to additional available constraints. Another possible approach is to completely determine $u$ and $u'$ (using methods such as interpolation, curve-fitting, or parameter estimation), and compute the distance on two completely specified utility functions. We believe that because of the strong assumptions required to compute the complete utility functions, the suitability of this approach can only be determined on a case-by-case basis. As we shall show in this paper, the probabilistic distance provides a principled solution for this problem that can be used in a wide range of other problems as well.

## 3  PRELIMINARY

In this section, we introduce the necessary background on orders, partial orders, value functions, utility functions, and utility theory. We will occasionally use the terms *decision alternative* and *decision consequence* interchangably, as we are mainly interested in the consequence of a decision.

### Complete Preference Orders

A *preference order* $\prec$ on a set of decision consequences $\mathcal{D}$ is a weak order, i.e. an asymmetric ($a \prec b \Rightarrow b \not\prec a$), negatively transitive ($a \not\prec b, b \not\prec c \Rightarrow a \not\prec c$) binary relation on $\mathcal{D}$. For $a, b \in \mathcal{D}$, $a \prec b$ indicates that the decision maker *prefers* $b$ to $a$. When neither of the two consequences is preferred to the other ($a \not\prec b, b \not\prec a$), we say that the decision maker is *indifferent* between them and denote this relation by $a \sim b$. An important technique that is often used in association with preference orders is the use of *consistent functions* that capture preference orders. A real-valued function $f : \mathcal{D} \to \Re$ is said to be *consistent* with a preference order $\prec$ on $\mathcal{D}$ if for all $a, b \in \mathcal{D}$, $a \prec b \Leftrightarrow f(a) < f(b)$. Any real-valued function $f : \mathcal{D} \to \Re$ induces a preference order $\prec_f$ according to the above $\Leftrightarrow$.

When the decision consequences are certain, we call them *outcomes*, and denote the set of outcomes by $\Omega$ (thus $\mathcal{D} = \Omega$). We will assume throughout the paper that $\Omega$ is finite and $\Omega = \{1, 2, \ldots, n\}$. It can be proven that for any preference order $\prec$ over $\Omega$ there exists a function $v$, called a *value function*, that is consistent with $\prec$. When the decision consequences are uncertain, they are modeled by probability distributions over outcomes and called *prospects*. We denote the set of all prospects, which is the set of all probability distributions over $\Omega$ by $\mathcal{S}$. The central result of utility theory is a representation theorem that identifies a set of conditions guaranteeing the existence of a func-tion consistent with the preference of a decision maker [17]. This theorem states that if the preference order of a decision maker satisfies a few "rational" properties, then there exists a real-valued function, called a *utility function* $u : \Omega \to \Re$, over outcomes such that $p \prec q \Leftrightarrow \langle p, u \rangle < \langle q, u \rangle$. Here $\langle p, u \rangle$, the inner product of the probability vector $p$ and the utility vector $u$, is the expected value of function $u$ with respect to the distribution $p$: $\langle p, u \rangle = E_p[u]$. It is often convenient to extend $u$, by means of expectation, to a function $u : \mathcal{S} \to \Re$ that maps a prospect $p \in \mathcal{S}$ to $\langle p, u \rangle$. This function is clearly consistent with the preference order $(\mathcal{S}, \prec)$. In this paper, we work only with preference orders that satisfy the above rational properties.

Two value (or utility) functions that induce identical preference orders are said to be *strategically equivalent*. (Note that strategic equivalence is an equivalence relation, denoted $\simeq$.) Otherwise, they are said to be *strategically different*.

**Partial Preference Orders**

How should one represent partial preferences? For the purpose of the case-based preference elicitation, a partial preference of a person is obtained via an incomplete elicitation, such as the one described in Section 2. For the most generality, we may assume that a partial preference order $\prec$ is a binary relation on the set $\mathcal{D}$ of decision consequences. Furthermore, it is reasonable to assume that this binary relation is asymmetric: if we know that a person prefers $a$ to $b$, then it is not the case that he prefers $b$ to $a$. We may also assume transitivity: if he prefers $a$ to $b$, and $b$ to $c$, then he prefers $a$ to $c$. In the theory of orders, an asymmetric, transitive binary relation is a called a *partial order*, or a *poset*. In this framework, we thus represent partial preferences using partial orders[1]. We have a slightly different concept of consistent functions for partial orders. A real-valued function $f : \mathcal{D} \to \Re$ over the decision consequences is said to be *consistent* with a partial preference order $\prec$ if for any decision consequences $a, b$, $a \prec b \Rightarrow f(a) < f(b)$ and $a \sim b \Rightarrow f(a) = f(b)$. The set of all functions that are consistent with $\prec$ is denoted as $C(\prec)$. Intuitively, consistent functions capture all information contained in the partial orders, and they might contain more than that. Consequently, functions that are consistent with a partial preference order $\prec$ may be strategically different, as they induce weak orders that are different extensions of $\prec$. There is however a one-to-one correspondence between the weak order extensions of $\prec$ and the equivalence classes of $(C(\prec), \simeq)$.

## 4 THE PROBABILISTIC DISTANCE ON COMPLETE PREFERENCES

In [8] we introduce the *probabilistic distance* as a measure of distance between two complete preference orders. Given two persons with corresponding (complete) preference orders $\prec_1$ and $\prec_2$, the probabilistic distance, denoted $\delta(\prec_1, \prec_2)$, is defined as the probability that a uniformly randomly chosen pair $(a, b)$ of decision consequences causes a *conflict* between the two users, i.e, the two users rank $a$ and $b$ differently. Formally, let the *conflict indicator function* $c_{\prec_1, \prec_2} : \mathcal{D}^2 \to \{0, 1\}$ be defined as follows:

$$c_{\prec_1, \prec_2}(a, b) := \begin{cases} 1 & \text{if } (a \preceq_1 b \wedge b \prec_2 a) \\ & \vee (a \prec_1 b \wedge b \preceq_2 a) \\ & \vee (a \preceq_2 b \wedge b \prec_1 a) \\ & \vee (a \prec_2 b \wedge b \preceq_1 a) \\ 0 & \text{otherwise.} \end{cases}$$

The probabilistic distance is formally defined as

$$\delta(\prec_1, \prec_2) := E[c_{\prec_1, \prec_2}(a, b)]. \tag{1}$$

Here the expectation is taken with respect to $a$ and $b$, which are two independent identically distributed uniform random variables on $\mathcal{D}$. The probabilistic distance is a metric on the set of preference orders: it is symmetric, and satisfies the triangle inequality and the "distinguishability of non-identicals" property [8].

### 4.1 THE CASE OF CERTAINTY

When the decision problem does not involve uncertainty, the distance $\delta(\prec_1, \prec_2)$ can be computed by simply averaging the conflict function $c_{\prec_1, \prec_2}(i, j)$ over all $n^2$ pairs $(i, j) \in \Omega^2$. Other popular metric on the set of permutations of $\{1, 2, \ldots, n\}$ include *Pearson's correlation coefficient*, *Spearman's rho*, *Spearman's footrule*, *Ulam's distance*. See Critchlow [5] for a discussion these metrics from a statistical point of view.

### 4.2 THE CASE OF UNCERTAINTY

In the case of certainty, defining and computing distance measures on preference orders seem relatively straightforward. Things get a little bit more complicated in the case of uncertainty. Let $\prec_1$ and $\prec_2$ be two preference orders on the set $\mathcal{S}$ of prospects.

---
[1] Note that the difference between the definition of complete preference order and that of partial preference order is the difference between negative transitivity and transitivity. Given asymmetry, transitivity is weaker than negative transitivity, i.e. the latter implies the former. This "weakness" reflects the incompleteness of our information about the person's preference.

Table 1: Algorithm for uniform sampling on $\mathcal{S}$.

1. Generate $n-1$ numbers $x_i, i = 1, 2, ..., n-1$ according to $n-1$ independent uniform random variables on $[0, 1]$.

2. Sort $x_i$'s: $0 \leq x_{(1)} \leq x_{(2)} \leq \ldots \leq x_{(n-1)} \leq 1$. This is the order statistics of the sample. Let $x_{(0)} = 0$ and $x_{(n)} = 1$.

3. Let $p_i = x_{(i)} - x_{(i-1)}, i = 1, 2, \ldots, n$. ($p_i$'s are called the spacings of the sample). Return $(p_1, p_2, \ldots, p_n)$.

The probabilistic distance is defined as $\delta(\prec_1, \prec_2) = E[c_{\prec_1, \prec_2}(p, q)]$, where $p$ and $q$ are two independent identically distributed uniform random variables on the set $\mathcal{D}$ of decision consequences. There are two possible ways to approach this definition. In the first approach, we assume no knowledge about the available decision alternatives and thus take $\mathcal{D}$ to be $\mathcal{S}$, the set of all prospects:

$$\delta(\prec_1, \prec_2) = \int_\mathcal{S} \int_\mathcal{S} c_{\prec_1, \prec_2}(p, q) \partial p \partial q. \qquad (2)$$

In previous work [8], we introduced this definition, but left open the non-trivial issue of how to compute the above integral. The main difficulty here is that while it can be noted that computing this integral amounts to computing the volume of a polytope in the $(2n - 2)$-dimension space (both $p$ and $q$ have $n - 1$ coordinates that can vary), the resulting volume-computing problem in general is computationally intractable [1]. There is, however, a simple Monte Carlo approximation algorithm for this particular problem. This algorithm works by sampling $p^{(i)}, i = 1, 2, \ldots, k$ and $q^{(i)}, i = 1, 2, \ldots, k$ according to the uniform distribution on $\mathcal{S}$, and taking the average $\bar{c} = \frac{1}{k} \sum_{i=1}^{k} c_{\prec_1, \prec_2}(p^{(i)}, q^{(i)})$. With a sufficiently big sample size $k$, the sample mean $\bar{c}$ can approximate $\delta(\prec_1, \prec_2)$ with arbitrary precision, according to the Central Limit Theorem. Sampling $p^{(i)}$ and $q^{(i)}$ according to the uniform distribution on $\mathcal{S}$ is basically the well-studied problem of *random division of the unit interval* and can be performed using the algorithm in Table 1 (see [14] for more details).

The probabilistic distance between two preference orders, defined this way, depends only on the orders. It can be computed given the two orders, or two utility functions that are consistent with the two orders. This definition can be useful when the two preference orders, or the two consistent utility functions are given, but little is known about the available decision alternatives. When we have more information about the decision alternatives and their consequences, it is desirable that we tailor the definition of the probabilistic distance to reflect this knowledge. So in the second approach to defining the probabilistic distance, we assume that the set $\mathcal{D}$ of decision consequences is finite and known. The probabilistic distance is defined as

$$\delta(\prec_1, \prec_2) = \frac{\sum_{(p,q) \in \mathcal{D}^2} c_{\prec_1, \prec_2}(p, q)}{|\mathcal{D}|^2}. \qquad (3)$$

The computation of this (discrete) formula is obviously much simpler than the integral formula of Equation 2, *provided that* we know the set of decision alternatives $\mathcal{D}$. Note that it is a subtle issue to determine which alternatives to include in $\mathcal{D}$ in the above definition.

## 5 PROBABILISTIC DISTANCE ON PARTIAL PREFERENCES

In [8], we proposed to extend the definition of probabilistic distance to partial orders in the following way. Let $\prec_1$ and $\prec_2$ be two partial orders with corresponding sets of weak order extensions $E_1$ and $E_2$. Recall that $E_i$ can be viewed as a set of strategically different value/utility functions $f_i$ consistent with $\prec_i$, for $i = 1, 2$. These functions form a one-to-one correspondence with the weak order extensions of $\prec_i$ (note that in the uncertainty case, the correspondence is with only extensions that satisfy the "rational properties" required for the existence of a utility function). We define the probabilistic distance $\delta(\prec_1, \prec_2)$ to be the average of the probabilistic distance between pairs of extensions of $\prec_1$ and $\prec_2$, respectively. Formally,

$$\begin{aligned}\delta(\prec_1, \prec_2) &= E[\delta(\prec_{f_1}, \prec_{f_2})] \\ &= E\left[E[c_{\prec_{f_1}, \prec_{f_2}}(a, b)]\right],\end{aligned}$$

where $f_i$ are uniform random variables on $E_i$, $i = 1, 2$, and $a$ and $b$ are uniform random variables on $\mathcal{D}$. Note that this distance is *not* a metric on the set of partial orders, since the distance between two identical partial orders that are not complete orders is always positive (which violates the "distinguishability of non-identicals" property). This, however, is desirable if the two orders represent the preferences of two different users, since the complete preference orders for the two may actually differ.

### 5.1 THE CERTAINTY CASE

In previous work [8], we have addressed the issue of computing the probabilistic distance on partial orders for the case of decision making under certainty. In this case, the set $\Omega$ of decision alternatives is finite, and so are the sets $E_1, E_2$ of weak order extensions of partial orders $\prec_1, \prec_2$. Thus, a simplistic approach would be to evaluate the conflict function c for all possible 4-tuples $\{(f_1, f_2, i, j) | f_1 \in E_1, f_2 \in E_2, i, j \in \Omega\}$ and take the average. This however is impractical because the number of weak order extensions of a partial order can be exponentially large (the number of strict order extensions of a vacuous partial order - a partial order in which everything is incomparable with everything - is $n!$). To get around this problem, we turned to an approximation approach. Instead of taking the (real) average of the conflict function $c$ for all $f_1 \in E_1$ and $f_2 \in E_2$, we use the Monte Carlo method and take the average of $c$ for only a sample set of $(f_1, f_2)$. This is made possible using an algorithm that samples almost uniformly randomly from $E_1$ and $E_2$. The sampling algorithm is due to Bubley and Dyer [2].

## 5.2 THE UNCERTAINTY CASE

In Section 4, we have seen that computing the probabilistic distance on complete orders become more complex when we go from the certainty to the uncertainty case. This suggests that the hardest issue of all is computing the probabilistic distance on *partial preference orders* in the *uncertainty case*. Let $\prec_1$ and $\prec_2$ be the partial preference orders of two persons, $A_1$ and $A_2$. Recall that the probabilistic distance $\delta(\prec_1, \prec_2)$ is defined as $\delta(\prec_1, \prec_2) = E[\delta(\prec_{f_1}, \prec_{f_2})]$, where $f_1, f_2$ are uniform random variables on $E_1, E_2$, the sets of weak order extensions of $\prec_1, \prec_2$, respectively. Exactly how should we interpret this definition? In the certainty case, this is easy since $E_1$ and $E_2$ are finite sets (a finite poset has only finitely many extensions) and we can just take the average of $\{\delta(\prec_{f_1}, \prec_{f_2}) | f_1 \in E_1, f_2 \in E_2\}$. But in the case of uncertainty, the set $E_1$ and $E_2$ are typically infinite. For example, consider a typical partial preference elicitation process. We may have determined that the utility function of $A_1$ is additive over two binary attributes $\{x_1, x_2\}$:

$$u(x) = k_1 u_1(x_1) + k_2 u_2(x_2), k_1, k_2 \geq 0. \quad (4)$$

In addition, we have also elicited the sub-utility functions $u_1, u_2$. We have not, however, assessed the scaling constants (or tradeoff coefficients) $k_1, k_2$. The set $E_1$ is thus the set of all utility functions of the form in Equation 4, which is obviously infinite.

**Partial Utility Functions As Polyhedral Cones**

Defining the expectation of a quantity involving random variables over infinite, multi-dimensional domains often requires the language and formalism of measure theory. With a simplifying assumption, however, we can define the probabilistic distance $\delta$ using more elementary concepts. Note that since a utility function $u : \Omega \to \Re$ can be viewed as a point in the $n$-dimensional Euclidean space $\Re^n$: $u = (u(1), u(2), \ldots, u(n))$, we can (and will) talk about the sets $E_1, E_2$ of consistent utility functions as sets of points in $\Re^n$. The simplifying assumption we shall make regarding $E_1, E_2$ is that they are determined by linear, homogeneous inequalities. Formally, they are sets of the forms

$$\{\vec{u} \in \Re^n | A\vec{u} \leq \vec{0}\}, \quad (5)$$

where $A$ is some $m \times n$ matrix of real numbers, and $\vec{0}$ is the $m \times 1$ zero vector. In geometric terms, such a set is the intersection of $m$ *half-spaces*, each of which crosses the origin and having one of the rows of matrix $A$ as its *outward normal vector*, and is called a *polyhedral cone*. Partial utility functions satisfying the above assumption encompass most of the common kinds of partial utility functions encountered in the practice of decision analysis. For example, a multi-linear utility function with known sub-utility functions and unknown scaling coefficients satisfies this assumption [9]. It is not difficult to see that the same is true for multiplicative and additive utility functions with known sub-utility functions and unknown scaling constants. Furthermore, a constraint on the partial preference order $\preceq$ of the form $p \preceq q$, for some $p, q \in \mathcal{S}$ would also translate to a homogeneous linear inequality: $\langle u, p - q \rangle \leq 0$.

The nice thing of having $E_1$ and $E_2$ as polyhedral cones is that in the defining formula of the probabilistic distance

$$\begin{aligned}\delta(\prec_1, \prec_2) &= E[\delta(\prec_{f_1}, \prec_{f_2})] \\ &= \int_{E_1} \int_{E_2} \int_{\mathcal{D}} \int_{\mathcal{D}} c_{\prec_{f_1}, \prec_{f_2}}(p, q) \partial f_1 \partial f_2 \partial p \partial q,\end{aligned}$$

we can interpret the integral on the right hand side as the *volume* of a bounded polyhedral cone in some multi-dimensional Euclidean space. But more importantly, we can reduce the problem of computing the probabilistic distance on partially specified utility functions to the well-studied problem of computing the volume of polyhedral cones. (In fact, the problem of computing the probabilistic distance on partial orders in the certainty case can also be reduced to the volume-computing problem, using some elementary geometric arguments.)

**Computing the Volume of Convex Bodies**

The problem of computing the volume of convex bodies has received considerable interest in the theoretical computer science community in the past fifteen years. Early results were negative for the prospect of finding an efficient deterministic algorithm [1]. But randomization techniques once again come to the rescue. The first work that uses randomization to obtain a polynomial time algorithm for this problem is due to Dyer et al [6]. A series of work followed and refined the algorithm of Dyer et al, substantially reducing its complexity [11]. These works are all based on various Markov chain-based sampling techniques that samples points from the convex body according to a nearly uniform distribution. The convex body is input to the algorithm by means of a *membership oracle*, i.e. a black box that provides the answer whether a given point belongs to the convex body. Note that this requirement fits excellently with the assumption that the set $E_1, E_2$ are polyhedral cones determined by a set of homogeneous linear inequalities as in Equation 5: we can check if a utility function $\vec{u}$ is consistent if $A\vec{u} \leq \vec{0}$ in time $O(m)$ (recall that $m$ is the number of rows of $A$).

In the rest of this section, we sketch out the main ideas behind the sampling algorithm. To sample uniformly from a convex body $K$, we perform a random walk on the points of $K$. Starting at an arbitrary point inside $K$, we move at each step to a uniformly selected random point in a ball of radius $\epsilon$ about the current point (if this remains inside $K$, if the new point is outside $K$, we remain where we were). The size $\epsilon$ of the radius is typically $1/\sqrt{n}$. It follows from elementary Markov chain theory that the distribution of the point after $t$ step tends to the uniform distribution as $t$ tends to infinity. The crucial issue is, how long to walk before the walking point becomes nearly uniformly distributed? There are two reasons for needing a long walk: we have to get to the "distant parts" of $K$, and we may get stuck in "corners", especially "sharp corner" of $K$. The first reason suggests that we choose a step-size that is large enough relative to the diameter of $K$, while the probability of the second can be reduced by choosing a small step-size. A number of advanced techniques have been developed to address this dilemma to ensure that the Markov chain settles quickly to a nearly uniform distribution (in technical terms, such a chain is called *rapidly mixing*). See Lovász et al [11] for a comprehensive treatment of this topic.

While this Markov chain-based sampling algorithm was developed for the purpose of computing the volume of convex bodies (and thus can be used to compute the volume of the polyhedron that is $\delta(\prec_1, \prec_2)$), we can use it directly to perform a Monte Carlo estimation of the probabilistic distance on partial utility functions. Specifically, we can estimate $\delta(\prec_1, \prec_2)$ by sampling $f_1^{(i)}, i = 1, 2, \ldots, k$ and $f_2^{(i)}, i = 1, 2, \ldots, k$ according to nearly uniform distributions on $E_1$ and $E_2$ respectively, and taking the average $\overline{\delta} = \frac{1}{k} \sum_{i=1}^{k} \delta(f_1^{(i)}, f_2^{(i)})$. Again, the Central Limit Theorem ensures that with a sufficiently big sample size $k$, the sample mean $\overline{\delta}$ can approximate $\delta(\prec_1, \prec_2)$ with arbitrary degree of precision.

## 6 AN ILLUSTRATIVE EXAMPLE

In this section, we illustrate the algorithm to compute the probabilistic distance on partially specified utility functions. The data we use are taken from the psychology experiment by Miyamoto and Eraker [12], as described in Section 2. Out of the 44 subjects, 6 were dropped due to failure to complete the interview in the allocated time, or failure to understand the CE task. The effective sample size is thus 38. There are a total of 42 CE questions (see Table 2). Note that with this data set, it is not possible to define a distance measure that requires the knowledge of the decision alternatives (Equation 3).

Since the survival duration in the CE questions ranges from 0 to 36, we scale the utility functions so that $u(0) = 0$ and $u(36) = 1$. The next step is to discretize the outcome space, which is discretizing the number of years of survival. Because each subject gave 4 different answers (at 4 different time points) to each CE questions, we take the average of the 4 answers as the CE. Because each answer is either integers or integers plus 0.5 (e.g. $(1, .5, 10) \sim 4.5$), we discretize the number of years of survival to the granularity of $1/8$, resulting in $36 \times 8 + 1 = 289$ outcomes. We also assume that all subjects prefer longer survival to shorter survival: $u(\frac{i}{8}) \leq u(\frac{i+1}{8}), i = 0, 1, \ldots, 287$. Framed this way, the utility function $u$ of each subject has a total of 288 inequality constraints and $42 + 2 = 44$ equality constraints. It is easy to see that these linear constraints determine a convex set of consistent utility functions.

To find a starting point for the random walk, we need to find a consistent utility function, i.e, a feasible solution for the linear constraints. For this we use the linear programming facility LINPROG of Matlab® Optimization Toolbox, with some randomly generated target function. Interestingly, we found that out of the 38 subjects, only 3 provided consistent answers; the rest provided answers that lead to linear programs that are infeasible. This inconsistency can be attributed to the fact that the expected utility paradigm is normative but not descriptive [10]. An example of this school of thought is the approach called *subjective expected utility* (SEU) [16], according to which a CE statement $(m, .5, n) \sim p$ translates into the equation: $(1 - w(.5))u(m) + w(.5)u(n) = u(p)$. Here $0 < w(.5) < 1$ is the *probability distortion* for a .5 probability applying to the superior outcome. Note that in the standard expected utility paradigm, $w(.5) = .5$.

But even with more general utility models such as SEU, it is likely that subjects will have inconsistent preferences, due to *variations in subject responses*.

Table 2: Standard gamble questions. $X/Y$ denotes a 50/50 gamble between $X$ and $Y$ years of survival.

| Basic | Times 2 | Times 3 | Plus 10 | Plus 20 | Zero |
|-------|---------|---------|---------|---------|------|
| 1/10  | 2/20    | 3/30    | 11/20   | 21/30   | 0/32 |
| 2/10  | 4/20    | 6/30    | 12/20   | 22/30   | 0/36 |
| 3/10  | 6/20    | 9/30    | 13/20   | 23/30   |      |
| 4/10  | 8/20    | 12/30   | 14/20   | 24/30   |      |
| 1/12  | 2/24    | 3/36    | 11/22   | 21/32   |      |
| 2/12  | 4/24    | 6/36    | 12/22   | 22/32   |      |
| 3/12  | 6/24    | 9/36    | 13/22   | 23/32   |      |
| 4/12  | 8/24    | 12/36   | 14/22   | 24/32   |      |

Our approach is to stay within the standard expected utility paradigm and account for the inconsistency in some way. While the fact that random error in judgement exists is well-known, the question of how to deal with it remains open. For the purpose of our experiment, we take the following simple approach. We keep all of the 288 inequality constraints that capture the "longer survival is better" assumption. For each subject, from the set of the 42 equality constraints provided by the CE answers, we incrementally randomly add one at a time to LINPROG and keep doing this as long as a feasible solution exists. Note that due to differences between subjects' responses and the randomness of this method, different sets of CE answers may be taken into account for different subjects. Fortunately, this is not a problem for the probabilistic distance.

Now that a set of consistent CE answers is selected for each subject, we simultaneously start 38 random walks from 38 consistent utility functions, one for each subject. The radius $\epsilon$ of the ball is initialized to 0.001. At each iterations, we generate a random point in each ball of radius $\epsilon$. If the generated point is consistent with the constraints, we move to the new point and mark the iteration a *success*; otherwise we stay at the current location and call the iteration a *failure*. If two successes occur consecutively, we double the radius. If two failures occur consecutively, we halve the radius. We stop the random walk after 1000 iterations, at which point we obtain a random sample of consistent utility functions for the 38 subjects. We compute the distance between any two consistent utility functions and record the distances in a square dissimilarity matrix of size $38 \times 38$. This computation is performed by a routine that implements the algorithm in Table 1. We repeat the whole process for a total of 1000 times, updating the averages of the distances as we go. Finally, we input the average distance matrix to the hierarchical clustering algorithm of ClustanGraphics® to obtain the hierarchical clustering shown in Figure 1. The method used was average-linkage. [2]

---

[2] All of the codes were written in Java$^{TM}$ and the math-

## 7 SUMMARY AND DISCUSSION

In previous work [8], we introduced the probabilistic distance as a measure of dissimilarity among people preferences, and provided algorithms to estimate this measure in the case of decision making under certainty. In this paper we complete the discussion of the probabilistic distance by providing algorithms to estimate this measure in the uncertainty case. Under uncetainty, the problem is innately harder, because of the complexity introduced by probabilities and utilities. We have shown that with the reasonable assumption that the set of consistent utility functions is linearly bounded, computing the probabilistic distance can be reduced to the well-studied problem of computing the volumes of convex bodies for which efficient approximate algorithms exist. A key ingredient of these algorithms is a Markov chain-based, polynomial time sampling algorithm that samples points from a convex body according to a nearly uniform distribution. We use this sampling algorithm directly to estimate the probabilistic distance on partially specified utility functions. We demonstrate this procedure on a set of partially specified utility functions elicited from 44 subjects who are undergraduates at the University of Michigan. We show how the probabilistic distance between subjects can be computed based on arbitrary sets of answers to standard gamble questions. Note that in computing the probabilistic distance, we can incorporate any prior knowledge about user utilities in the form of utility constraints, as long as the constraints are linear. The more constraints there are, the more accurately the distance measure can be computed. To our knowledge, this work is the first attempt to define a similarity measure on partial utility functions and to develop a method to compute this measure. The implication of the probabilistic distance goes beyond the context of case-based preference elicitation, since it is in its most general form a *distance measure on partial orders* - a topic that has not been received adequate treatment.

We are currently investigating several medical decision problems as potential candidates for implementing the case-based preference elicitation approach. For such candidates, the basic requirement is that a database of patient utilities is available. Since utility data are routinely collected for a wide range of medical decision problems, and since the standard gamble CE method is one of the most widely used techniques to elicit utilities, we believe that the case-based approach using the probabilistic distance has serious potential to see

---

ematical programming language of MatLab®. The computations were performed on an Athlon$^{TM}$@850Mhz system with 512MB RAM running Windows® 2000, and took about an 30 minutes to finish.

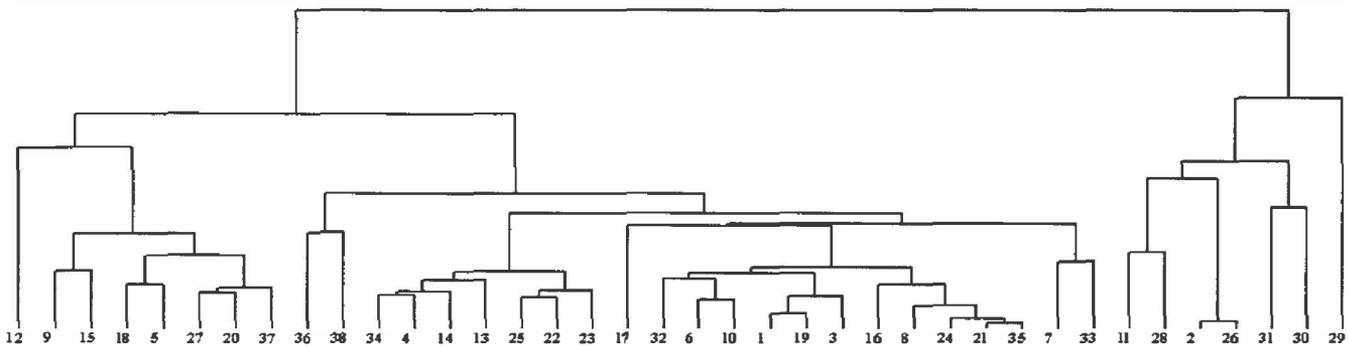

Figure 1: Hierarchical cluster of the 38 subjects.

real-world application.

Chajeswska et al. [3] pursue an approach to utility elicitation that is somewhat similar to ours. They also start from an assumption that there exists a database of utility functions, partially or completely specified. This assumption differs from ours in that here the database needs to contain the *actual utilities* (as opposed to *constraints* on utilities). The novelty of this approach is that utilities are treated as random variables, and if drawn from a mixture of Gaussians, as they were postulated to, their density functions can be learned from the utility database using Bayesian learning techniques. Also, using standard Bayesian techniques, it is possible to determine the relevance of an elicitation question based on its *value of information* [4]. In contrast, our case-based approach requires fewer structural assumptions and as such has an edge over Chajewska et al.'s approach in those situations where these assumptions are not applicable.